\pdfoutput=1

\documentclass[11pt]{article}

\usepackage{emnlp2021}

\usepackage{times}
\usepackage{latexsym}

\usepackage{amsmath,amssymb}
\usepackage{xspace}
\usepackage{booktabs}
\usepackage{graphicx}
\usepackage{multirow}
\usepackage{xcolor,colortbl}
\usepackage{subcaption}
\usepackage[T1]{fontenc}
\usepackage[utf8]{inputenc}

\usepackage{microtype}

\newcommand{\fscore}[1][1]{$F_{#1}$\xspace} 

\newcommand{\corpusIdNerCONLL}{\textsc{Conll}\xspace}
\newcommand{\corpusIdNerClinical}{\textsc{i2b2-Clin}\xspace}
\newcommand{\corpusIdNerAnonymization}{\textsc{i2b2-Anon}\xspace}
\newcommand{\corpusIdNerTwitter}{\textsc{Wnut-16}\xspace}
\newcommand{\corpusIdNerSocial}{\textsc{Wnut-17}\xspace}
\newcommand{\corpusIdNerWetlab}{\textsc{Wnut-20}\xspace}
\newcommand{\corpusIdNerLitbank}{\textsc{LitBank}\xspace}
\newcommand{\corpusIdNerSEC}{\textsc{Sec}\xspace}
\newcommand{\corpusIdNerSOFC}{\textsc{Sofc}\xspace}

\newcommand{\corpusIdGum}{\textsc{Gum-All}\xspace}
\newcommand{\corpusIdGumAcademic}{\textsc{Gum-Acad}\xspace}
\newcommand{\corpusIdGumBiography}{\textsc{Gum-Bio}\xspace}
\newcommand{\corpusIdGumFiction}{\textsc{Gum-Fict}\xspace}
\newcommand{\corpusIdGumInterview}{\textsc{Gum-Int}\xspace}
\newcommand{\corpusIdGumNews}{\textsc{Gum-News}\xspace}
\newcommand{\corpusIdGumReddit}{\textsc{Gum-Red}\xspace}
\newcommand{\corpusIdGumTravel}{\textsc{Gum-Trav}\xspace}
\newcommand{\corpusIdGumWikihow}{\textsc{Gum-Whow}\xspace}

\newcommand{\corpusIdPosEWT}{\textsc{Ewt}\xspace}

\newcommand{\corpusIdPosLines}{\textsc{LinEs}\xspace}
\newcommand{\corpusIdPosPartut}{\textsc{ParTut}\xspace}

\newcommand{\corpusIdTimeTimebank}{\textsc{TimeBank}\xspace}
\newcommand{\corpusIdTimeAquaint}{\textsc{Aquaint}\xspace}
\newcommand{\corpusIdTimeAncient}{\textsc{Ancient}\xspace}
\newcommand{\corpusIdTimeWikiwars}{\textsc{Wwars}\xspace}
\newcommand{\corpusIdTimeSMS}{\textsc{Time4sms}\xspace}
\newcommand{\corpusIdTimeSCI}{\textsc{Time4sci}\xspace}
\newcommand{\corpusIdTimeClinical}{\textsc{i2b2-Time}\xspace}
\newcommand{\corpusIdAce}{\textsc{Ace-All}\xspace}
\newcommand{\corpusIdAceNews}{\textsc{Ace-Bn}\xspace}
\newcommand{\corpusIdAceConversations}{\textsc{Ace-Bc}\xspace}
\newcommand{\corpusIdAceTelephony}{\textsc{Ace-Cts}\xspace}
\newcommand{\corpusIdAceNewswire}{\textsc{Ace-Nw}\xspace}
\newcommand{\corpusIdAceUsenet}{\textsc{Ace-Un}\xspace}
\newcommand{\corpusIdAceWebblog}{\textsc{Ace-Wb}\xspace}

\newcolumntype{a}{>{\columncolor{blue!5}}l}
\newcolumntype{b}{>{\columncolor{blue!5}}c}
\newcommand{\best}[1]{\textbf{\textbf{#1}}}

\newcommand{\avgRank}{$\rho$\xspace}
\newcommand{\knn}[1]{${#1}$-NN\xspace}

\newcommand{\topn}[1]{\textit{Top-}${#1}$\xspace}

\title{To Share or not to Share: \\Predicting Sets of Sources for Model Transfer Learning}

\author{Lukas Lange$^{1,2,3}$ \\
	\And
	Jannik Str\"{o}tgen$^1$ \\
	\hspace{4cm}$^1$ Bosch Center for Artificial Intelligence, Renningen, Germany\\
	\hspace{4cm}$^2$ Spoken Language Systems (LSV), Saarland University, Saarbr\"{u}cken, Germany\\
	\hspace{4cm}$^3$ Saarbr\"{u}cken Graduate School of Computer Science, Saarbr\"{u}cken, Germany\\
	{\tt \hspace{4cm}\{Lukas.Lange,Heike.Adel,Jannik.Stroetgen\}@de.bosch.com} \\
	{\tt \hspace{4cm}dietrich.klakow@lsv.uni-saarland.de} \\
	\And
	Heike Adel$^1$ \\
	\And
	Dietrich Klakow$^2$ \\
	\\}

\begin{document}
\maketitle

\begin{abstract}
In low-resource settings, model transfer can help to overcome a lack of labeled data for many tasks and domains. However, predicting useful transfer sources is a challenging problem, as even the most similar sources might lead to unexpected negative transfer results. 
Thus, ranking methods based on task and text similarity --- as suggested in prior work --- may not be sufficient to identify promising sources.
To tackle this problem, we propose a new approach to automatically determine which and how many sources should be exploited.
For this, we study the effects of model transfer on sequence labeling across various domains and tasks and show that our methods based on model similarity and support vector machines are able to predict promising sources, resulting in performance increases of up to 24 \fscore points.
\end{abstract}

\section{Introduction}
For many natural language processing applications in non-standard domains, only little labeled data is available. This even holds for high-resource languages like English \cite{klie-etal-2020-zero}.
The most popular method to overcome this lack of supervision is transfer learning from high-resource tasks or domains. 
This includes the usage of resources from similar domains \cite{ruder-plank-2017-learning}, domain-specific pretraining on unlabeled text \cite{gururangan-etal-2020-dont}, and the transfer of trained models to a new domain \cite{bingel-sogaard-2017-identifying}.
While having the choice among various transfer sources can be advantageous, 
it becomes more challenging to identify the most valuable ones
as many sources might 
lead to negative transfer results, i.e., actually reduce performance \cite{pruksachatkun-etal-2020-intermediate}.

\begin{figure}
    \centering
    \includegraphics[trim=5 5 0 0,clip,width=0.41\textwidth]{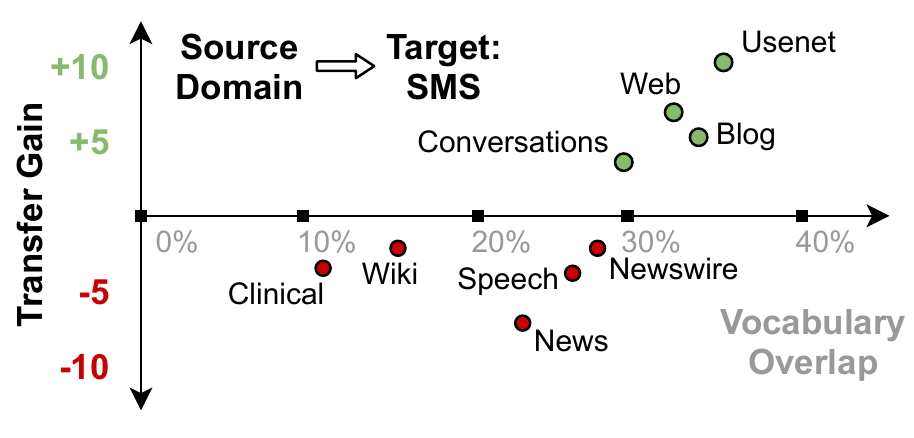}
    \caption{Observed transfer gains by transferring models from a source corpus to SMS texts. The domains are sorted by their vocabulary overlap to the target. 
    }
    \label{fig:intro}
\end{figure}

Current methods to select transfer sources are based on text or task similarity measures \cite{dai-etal-2019-using,schroder-biemann-2020-estimating}. The underlying assumption is that similar texts and tasks can support each other. 
An example for similarity based on vocabulary overlap is shown in Figure~\ref{fig:intro}.
However, current methods typically consider text and task similarity in isolation, which limits their application in transfer settings where both the task and the text domain change.

Thus, as a first major contribution, this paper proposes a new model similarity measure that represents text and task similarity jointly. 
By learning a mapping between two neural models, it captures similarity between domain-specific models across tasks. 
We perform experiments for different transfer settings, namely zero-shot model transfer, supervised domain adaptation and cross-task transfer across a large set of domains and tasks. Our newly proposed similarity measure successfully predicts the best transfer sources and outperforms existing text and task similarity measures.

As a second major contribution,
we introduce a new method to automatically
determine which and how many sources should be used in the transfer process, as the transfer can benefit from multiple sources. Our selection method overcomes the limitations of current transfer methods, which solely predict single sources based on rankings.
We show the benefits of transfer from sets of sources and demonstrate that support vector machines are able to predict the best sources 
across domains and tasks.
This improves performance with absolute gains of up to 24 \fscore points and effectively prevents negative transfer. 

The code for our sequence taggers, prediction methods and the results are publicly available.\footnote{\url{https://github.com/boschresearch/predicting_sets_of_sources}}

\section{Related Work}

\textbf{Domain adaptation \& transfer learning}
are typically performed by transferring information and knowledge from a high-resource to a low-resource domain or task 
\cite{daume-iii-2007-frustratingly,Ruder2019Neural, hedderich-2021-survey}. 
Recent approaches can be divided into two groups: (i)
model transfer \cite{ruder-plank-2017-learning} by reusing trained task-specific weights \cite{vu-etal-2020-exploring,lange-etal-2021-meddoprof} or by first adapting models on the target domain before training the downstream task \cite{gururangan-etal-2020-dont,rietzler-etal-2020-adapt} and
(ii) multi-task training \cite{collobert2008unified} where multiple tasks are trained jointly 
by learning shared representations \cite{peng-dredze-2017-multi,meftah-etal-2020-multi}.\
We follow the first approach in this paper.

For transfer learning, the \textbf{selection of sources} is utterly important.  
Text and task similarity measures \cite{ruder-plank-2017-learning,bingel-sogaard-2017-identifying} 
are used to select the best sources for cross-task transfer \cite{jiang-etal-2020-generalizing}, multi-task transfer \cite{schroder-biemann-2020-estimating}, cross-lingual transfer \cite{chen-etal-2019-multi} and language modeling \cite{dai-etal-2019-using}.
Alternatively, neural embeddings for corpora can be compared \cite{vu-etal-2020-exploring}.
In prior work, the set of domains is usually limited and the focus is on the single-best source. In contrast, we
exploit sources from a larger set of domains
and also explore the prediction of sets of sources, 
as using 
multiple sources is likely to be beneficial, 
as also shown by \newcite{parvez2021evaluating} contemporaneously to this work.

\section{Similarity Measures and Predictors}
In this section, we describe the sequencer tagger model and similarity measures along with metrics for the evaluation. Finally, we  introduce our new prediction method for sets of transfer sources.

\subsection{Terminology}
We consider two dimensions of datasets: the task $T$, which defines the label set, and the input text coming from a specific domain $D$. We thus define a dataset as a tuple $\langle T, D\rangle$, and specify in our experiments which of the two dimensions are changed.

\subsection{Similarity Measures}
\label{sec:similarity}
We apply the following measures to rank  
sources according to their similarity with the target data.

\textbf{Baselines.} We use the most promising domain similarity measures reported by \citet{dai-etal-2020-selection}: 
\textit{Vocabulary} and
\textit{Annotation overlap},
\textit{Language model perplexity} \cite{baldwin-etal-2013-noisy}, \textit{Dataset size} \cite{bingel-sogaard-2017-identifying} and \textit{Term distribution} \cite{ruder-plank-2017-learning}.
We also compare to 
domain similarity via \textit{Text embedding}s 
and task similarity using \textit{Task embedding}s \cite{vu-etal-2020-exploring}.

\textbf{Model similarity.}
As a new strong method, we propose \textit{Model similarity} that is able to combine domain and task similarity. 
For this, feature vectors $f$ for a target dataset $t$ are extracted from the last layer of two models $m_s, m_t$ which have been trained on the source and target datasets, respectively.
The features are then aligned by a linear transformation $W$, a learned mapping, between the feature spaces using the Procrustes method \cite{schonemann1966generalized} 
to minimize their pointwise differences:
\begin{equation}
    \textrm{arg min}_{W} | W(f(m_s, t))  - f(m_t, t) |
\end{equation}

The resulting transformation $W$ is the optimal mapping between the features $f(m_s, t)$ to $f(m_t, t)$. If both feature spaces are the same, $W$ would be the identity matrix $I$, i.e., no change is required for the transformation. 
Larger changes indicate dissimilarity, thus the distance between the two models is the difference of the mapping $W$ and the identity matrix $I$:
$\textrm{diff}(m_s, m_t) := | W - I |$.

Similar mappings have been used for the alignment of different embedding spaces \cite{emb/mikolov2013efficient,artetxe2018generalizing} as they inherently carry information on the relatedness between models.

\subsection{Prediction Methods for Sets of Sources}\label{sub:knn}

While these similarity measures can be applied to create rankings and select similar datasets, they still have a major shortcoming in practice:
None of them provides explicit insights when positive or negative transfer 
can be expected. 

Typically, the most similar source is selected for training based on a given similarity measure. 
We call this method \topn{1}.
This might introduce only a low risk of selecting a negative transfer source, but it also cannot benefit from further positive transfer sources. 
Thus, we also test its extension to an arbitrary selection of the $n$ best sources denoted by \topn{n}.
However, it is unclear how to choose $n$, and increasing $n$ comes with the risk of including sources that lead to negative transfer results.

As a solution, we propose two methods that predict whether positive transfer is likely for a given distance between datasets:
The first method models the prediction as a 3-class classification task, and the second one as a regression task predicting the transfer gain. 
For classification, we split the transfer gain $g$ into the three classes positive ($g \geq \theta$), neutral ($|g| < \theta$) and negative ($g \leq -\theta$) based on a predefined threshold $\theta$.  (In our experiments, we set $\theta=0.5$.)
We introduce the neutral class for classification to cope with small transfer gains $|g| < \theta$ that do not provide additional information, but would increase the training time.

To solve these tasks, we propose to use support-vector machines (SVM) for classification (-C) and regression (-R) and compare to $k$-nearest-neighbour classifiers (\knn{k}) as well as logistic and linear regression in our experiments.\footnote{We use sklearn implementations \cite{scikit-learn}.}
For each method, the input to the model is a similarity value between source and target. The training label is either the observed transfer gain (for regression) or the corresponding class (for classification) for the source-target pair. 
Given a new similarity value, a trained model can then be used to predict which kind of transfer can be expected.\footnote{
Other similarity measures can be included by modeling each value as a different input dimension. However, we found no significant improvements by including multiple measures.}  
The predictions for a target and a set of sources can then be used to select the subset of sources with expected positive transfer. 

\section{Experimental Setup}
In this section, we introduce the tasks, datasets and transfer settings used in our experiments.

\subsection{Tasks and Evaluation Metrics}
We perform experiments on 33 datasets for three tasks: Named entity recognition (NER), part-of-speech tagging (POS), and temporal expression extraction (TIME). 

For TIME tagging and for POS tagging, we use the English corpora described by \citet{strotgen2016domain} and the four publicly available universal dependencies corpora with the UPOS tag \cite{nivre-etal-2016-universal}, respectively. Following \newcite{lange-etal-2020-adversarial}, we convert the TIMEX corpora into the BIO format for sequence tagging. 
For NER with different label sets, we collected several datasets from a wide range of domains, including clinical \cite[\textsc{i2b2,}][]{corpus/i2b2/stubs2015}, social media \cite[\textsc{Wnut,}][]{strauss-etal-2016-results} and materials science corpora \cite[\textsc{Sofc,}][]{friedrich-etal-2020-sofc}.
The \textsc{Gum} \cite{corpus/zeldes-2017-gum} and \textsc{Ace}'05  \cite{corpus/walker-etal-2006-ace} can be split easily into multiple domains. Thus, we perform experiments for all subcorpora.
The \textsc{Gum} corpus has multi-layer annotations and includes named entity annotations as well. We use this to study the effects of NER transfer when the label set is shared.
All datasets are listed in the appendix with information on their domain and size with respect to the label set and number of sentences in the training, development, and test splits.

The metric for all experiments is micro \fscore. 
We use the difference in \fscore to measure transfer effects and also report \textit{transfer gain} \cite{vu-etal-2020-exploring}, i.e., the relative improvement of a transferred model compared to the single-task performance. 

In Section \ref{sec:ranking}, we rank sources according to their similarity to the target. These rankings are evaluated with two metrics, following \newcite{vu-etal-2020-exploring}: (1) the average rank of the best performing model in the predicted ranking denoted by \avgRank 
and (2) the normalized discounted cumulative gain \cite[NDCG,][]{javerlin-kekalainen-2002-ndcg}. 
The latter is a ranking measure commonly used in information retrieval, which 
evaluates the complete ranking while \avgRank only considers the top.

\subsection{Sequence Tagger Model}
For sequence tagging, we follow \citet{devlin-etal-2019-bert} and use BERT-base-cased as the feature extractor
and a linear mapping to the label space followed by a softmax as the classifier.

Models are trained using the AdamW optimizer \cite{loshchilov-hutter-2019-decoupled} with a learning rate of $2e-5$. The training is performed for a maximum of 100 epochs. We apply early stopping after 5 epochs without change of the \fscore-score on the development set. 
We use the same hyperparameters across all settings.\footnote{All our experiments are run on a carbon-neutral GPU cluster. The training of a single model takes between 5 minutes and 8 hours depending on the dataset size on a single Nvidia Tesla V100 GPU with 32GB VRAM.}

\setlength\tabcolsep{2.7pt}
\begin{table}
\footnotesize
\centering
\begin{tabular}{l|ll|ll|ll} \toprule
Task & \multicolumn{2}{c|}{Min.} & \multicolumn{2}{c|}{Avg.} & \multicolumn{2}{c}{Max.} \\ \midrule
\multicolumn{7}{l}{\textit{Zero-shot Model Transfer}} \\
\enspace NER  & -57.3 & (-37.9) & -17.7 & (-10.1) & 18.1 & (8.0) 
\\
\enspace POS  & -8.7 &  (-8.4)  & -2.8 & (-2.7)  &  1.6 & (1.5) 
  \\
\enspace TIME & -100.0 & (-83.2) & -42.7 & (-29.6) & 38.6 & (13.7)   
\\
\multicolumn{4}{l}{\textit{Supervised Domain Adaptation}} \\
\enspace NER  & -5.2 & (-2.7)   & 3.8 & (1.9) & 14.5 & (6.3) 
  \\
\enspace POS  & -0.3 & (-0.3)   & 0.4 & (0.4) & 1.8 & (1.7)
  \\
\enspace TIME & -15.3 & (-10.1) & 3.4 & (2.0) & 32.7 & (15.1)
  \\
\multicolumn{4}{l}{\textit{Cross-Task Transfer}} \\
\enspace NER$\rightarrow$NER  & -9.1 & (-4.1) & -0.2 & (-0.2) & 6.8 & (3.1) 
  \\
\enspace POS$\rightarrow$NER  & -5.9 & (-4.8) & -0.5 & (-0.3) & 2.6 & (1.2) 
  \\
\enspace TIME$\rightarrow$NER & -7.2 & (-3.3) & -0.9 & (-0.5) & 0.9 & (0.6) 
  \\ \midrule
\end{tabular}
\caption{Statistics on transfer gains (\fscore differences) for the three transfer settings. The average is aggregated over all domains and the 5 random seeds resulting in 210 task-specific experiments for NER and up to 780 for TIME. }
\label{tab:gains-single-task}
\end{table}
\setlength\tabcolsep{6pt}

\subsection{Transfer Settings}

\textbf{Zero-shot model transfer.}
We apply a model trained on a source dataset to a target with the same task but a different domain: $\langle T_i, D_i\rangle \rightarrow \langle T_i, D_j\rangle$. 

\textbf{Supervised domain adaptation.}
A model trained on a source domain is adapted to a target domain by finetuning its weights on target training data: $\langle T_i, D_i\rangle \rightarrow \langle T_i, D_j\rangle$.

\textbf{Cross-task transfer.}
For applying a model to a different task, we replace the classification layer with
a randomly initialized layer 
and adapt it to the new target task: $\langle T_i, D_i\rangle \rightarrow \langle T_j, D_j\rangle$.\footnote{We restrict the cross-task transfer to NER 
targets with different label sets,
as the combination of all tasks quickly becomes computationally infeasible given the large number of different settings.}

\section{Results}\label{sec:results}
This section presents the results of the different transfer settings and analyzes how similarity measures can be used to predict 
transfer sources.

\subsection{Analysis of Transfer Performance}
Table~\ref{tab:gains-single-task} shows the observed performance gains compared to the single-task performance. 
For zero-shot model transfer, we observe severe performance drops. 
In addition to domain-specific challenges, this setting is impaired by differences in the underlying annotation schemes.\footnote{For example, the \textsc{Timex2} \cite{ferro2005standard} and \textsc{Timex3} \cite{pustejovsky2005specification} guidelines disagree about including preceding words in the annotated mentions as "in".}

Supervised domain adaptation, i.e., adapting a model to the target domain, improves performance across all settings independent of the source domain. Table \ref{tab:gains-single-task} shows that the average transfer gains are positive for all tasks and that the maximum transfer gain is 32.7 
points for TIME.

The gains for cross-task transfer are smaller than for supervised domain adaptation. While we still observe some performance increases, the average transfer gains are negative for all tasks. This shows that it is likely that the adaptation of models from other tasks will decrease performance.
These results demonstrate the need for reliable similarity measures and methods to predict the expected transfer gains given the source task and domain. 
We will explore them in Section \ref{sec:ranking} and Section \ref{sec:many}.

\setlength\tabcolsep{3.3pt}
\begin{table}
\centering
\footnotesize
\begin{tabular}{l|cc|cc|cc|cc} \toprule
 & \multicolumn{2}{c|}{\begin{tabular}[c]{@{}l@{}}Model \\ Transfer\end{tabular}} & \multicolumn{2}{c|}{\begin{tabular}[c]{@{}l@{}}Domain \\ Adapt.\end{tabular}} & \multicolumn{2}{c|}{\begin{tabular}[c]{@{}l@{}}Cross \\ -Task\end{tabular}} & \multicolumn{2}{c}{Avg.} \\
Distance & \avgRank & N & \avgRank & N & \avgRank & N & \avgRank & N \\ \midrule
Vocabulary & \textbf{2.4} & \textbf{92.1} & \textbf{2.8} & 88.9 & 6.4 & 84.9 & 3.9 & 88.7 \\
Annotation & \textbf{2.4} & 91.7 & 3.1 & \textbf{89.3} & 6.1 & 85.3 & 3.9 & \textbf{89.1} \\
Dataset size & 3.6 & 86.4 & 3.8 & 85.9 & 7.2 & 82.3 & 4.9 & 84.9 \\
Term Dist & 2.8 & 90.5 & 4.2 & 87.5 & 6.7 & 85.2 & 4.5 & 87.7 \\ \midrule
LM Perp. & 3.9 & 85.6 & 3.4 & 88.2 & 5.9 & 84.4 & 4.4 & 86.1 \\
Text Emb. & 4.0 & 88.1 & 4.6 & 85.0 & 7.1 & 84.6 & 5.2 & 85.9 \\
Task Emb. & 4.1 & 88.5 & 4.7 & 84.8 & 6.6 & 84.5 & 5.1 & 85.6 \\
Model Sim. & 2.8 & 90.8 & 3.3 & 88.7 & \textbf{5.1} & \textbf{85.4} & \textbf{3.7} & 88.3 \\ \bottomrule
\end{tabular}
\caption{Ranking results for different similarity measures in the three transfer settings. Corpus-based measures are listed first and model-based ones below. The values displayed are the average rank of the best model (\avgRank) and the NDCG-score (N).}
\label{tab:predictions2}
\end{table}
\setlength\tabcolsep{6pt}

\begin{table*}[t!]
  \scriptsize
  \centering
\begin{tabular}{llll} \toprule
Target & Method & Sources & \fscore (increase) \\ \midrule
\multirow{10}{*}{\rotatebox{90}{TIME - \corpusIdAceUsenet}} & Single-Task Performance & \textit{No source corpora for pretraining} & 60.5 \fscore \\
 & Top-1 & \corpusIdTimeWikiwars & + 10.9 \\
 & Top-2 & \corpusIdTimeWikiwars, \corpusIdAceNews & + 11.2 \\
 & Top-3 & \corpusIdTimeWikiwars, \corpusIdAceNews, \corpusIdTimeTimebank & + 15.8 \\
 & All & \corpusIdTimeWikiwars, \corpusIdAce, \corpusIdTimeTimebank, \corpusIdTimeAquaint, \corpusIdTimeClinical, \corpusIdTimeAncient, \corpusIdTimeSCI, \corpusIdTimeSMS & + 10.2 \\
 & SVM (Classifier) & \corpusIdTimeWikiwars, \corpusIdAce, \corpusIdTimeTimebank, \corpusIdTimeAquaint, \corpusIdTimeSCI, \corpusIdTimeSMS & \textbf{+ 24.0}\\
 & Logistic Regression & \corpusIdTimeWikiwars, \corpusIdAce, \corpusIdTimeTimebank, \corpusIdTimeAquaint, \corpusIdTimeAncient, \corpusIdTimeSCI, \corpusIdTimeSMS & + 18.2 \\
 & k-Nearest-Neighbor & \corpusIdTimeWikiwars, \corpusIdAce, \corpusIdTimeTimebank, \corpusIdTimeAquaint, \corpusIdTimeClinical, \corpusIdTimeSCI, \corpusIdTimeSMS & + 17.1 \\
 & SVM (Regression) & \corpusIdTimeWikiwars, \corpusIdAce, \corpusIdTimeTimebank, \corpusIdTimeAquaint, \corpusIdTimeSCI, \corpusIdTimeSMS & \textbf{+ 24.0} \\
 & Linear Regression & \corpusIdTimeWikiwars, \corpusIdAce, \corpusIdTimeTimebank, \corpusIdTimeAquaint, \corpusIdTimeAncient, \corpusIdTimeSCI, \corpusIdTimeSMS & + 18.2 \\

\bottomrule
\end{tabular}
\caption{Predicted transfer sources for TIME domain adaptation with target \corpusIdAceUsenet (usenet).}
\label{tab:pred-src-time}
\end{table*}

\subsection{Similarity-based Ranking}\label{sec:ranking}
To evaluate the prospects of different sources for model transfer, 
we compute the pairwise distances between all datasets using the similarity measures presented in Section \ref{sec:similarity} and rank them accordingly.

Table~\ref{tab:predictions2} shows that the text-based methods vocabulary and annotation overlap are most suited for in-task transfer, i.e., model transfer and domain adaptation, while our model similarity is most useful for cross-task transfer. This shows that task similarity alone is not the most decisive factor for predicting promising transfer sources and domain similarity is equally or even more important, in particular, when more distant domains are considered. 
Our model similarity is able to capture both properties. 
It is the best similarity measure on average across all transfer settings according to the predicted rank of the top-performing source (\avgRank) and the best neural method according to NDCG. 
A more fine-grained analysis is given in Table~\ref{tab:predictions} in the appendix. 

In general, we find that selecting only the top source(s) based on a ranking from a distance measure, as done in current research, gives no information on whether to expect positive transfer.
Thus, we now explore methods to automatically predict sets of promising sources.

\subsection{Prediction of Sets of Sources}\label{sec:many}

\begin{figure}[t!]
    \centering
    \includegraphics[trim=0 10 0 0,clip,width=0.48\textwidth]{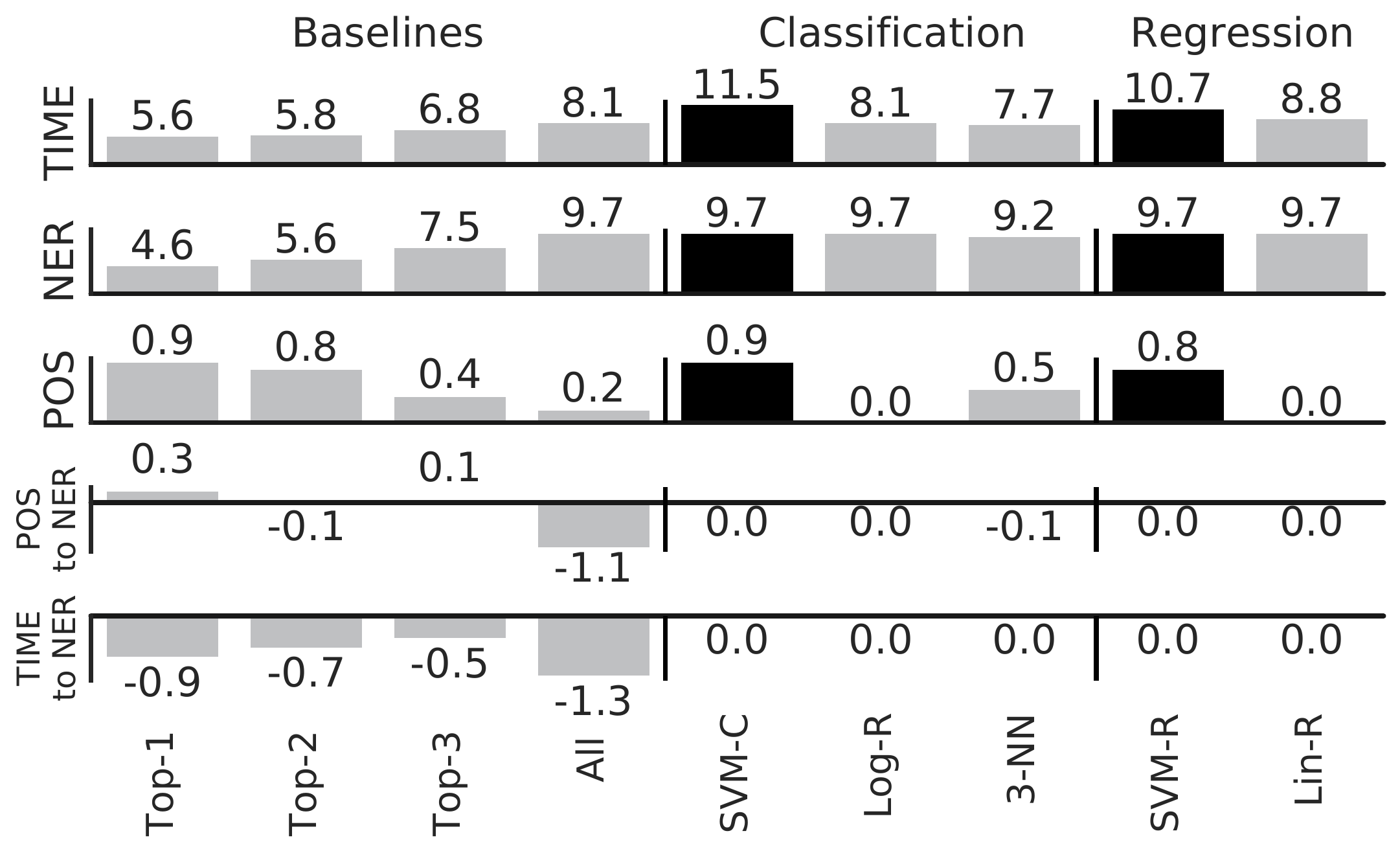}
    \caption{Average transfer gains using different classifiers for predicting the set of most promising sources.}
    \label{fig:performance-predictors}
\end{figure}

We use the methods introduced in Section~\ref{sub:knn} to predict the set of most promising sources. Then, we train a model on the combination of the selected sources and adapt it to the target.\footnote{We do not explore the NER to NER setting, as we restrict the sources to have the same set of labels.
For the other tasks, we trained source combinations which were predicted by at least one model (SVM-R/C, Log-/Lin-R, k-NN) or baseline method (Top-1, 2, .., All). Training all possible combinations would be be infeasible. }

The results averaged across the different settings are visualized in Figure~\ref{fig:performance-predictors}. 
While NER and TIME targets benefit from training on many sources, POS tagging targets gain the most from using only one or two of the most related source domains.
We find that our methods based on SVMs are able to predict this behavior and assign fewer sources for POS targets, and more sources for TIME and NER settings. In particular, for TIME settings, our methods SVM-C and -R result in much higher transfer gains compared to the static ranking-based methods and other classifiers or regression models. 

For example, transferring multiple sources using our SVM classifier to the \corpusIdAceUsenet target (see Table~\ref{tab:pred-src-time}) increases performance from 60.5 \fscore for single-task training to 84.5 \fscore (+24.0), which is much higher than the 10.9 points increase when using the single best source or 10.2 points using all available sources. 

For the cross-task experiments in the lower part of Figure~\ref{fig:performance-predictors}, we find that even the inclusion of the single best-ranked model results in a transfer loss of -0.9 points on average for TIME$\rightarrow$NER. In this setting, our models correctly adapt to this new challenge and predict an empty set of sources, indicating that no transfer should be performed.

\section{Conclusion}
We explored different transfer settings across three sequence-labeling tasks and various domains. 
Our new model similarity measure based on feature mappings outperforms currently used similarity measures as it is able to capture both task and domain similarity at the same time. 
We further addressed the automatic selection of sets of sources as well as the challenge of potential negative transfer by proposing a selection method based on support vector machines.
Our method results in performance gains of up to 24 \fscore points. 

\bibliography{references}
\bibliographystyle{acl_natbib}

\appendix

\section{Datasets}
All datasets are listed in Table~\ref{tab:datasets} with information on their domain and size with respect to the label set and number of sentences. We take the last 20\% and 10\% of the training data as test or development data whenever no split was provided.

\begin{table*}[thp!]
\footnotesize
\centering
\begin{tabular}{lllcc} \toprule
Task & Corpus & Domain
& \# Labels & \begin{tabular}[c]{@{}c@{}}\# Train / Dev / Test\\ sentences\end{tabular} \\ \midrule
\multirow{9}{*}{\rotatebox{90}{NER}} 
 & \multicolumn{2}{l}{\corpusIdNerCONLL \cite{corpus/conll/Sang03} News}   &  9 & 14,987 / 3,466 / 3,684 \\
 & \corpusIdNerClinical \cite{corpus/i2b2/uzuner2010} & Clinical concepts              &  7 & 13,052 / 3,263 / 27,625 \\
 & \corpusIdNerAnonymization  \cite{corpus/i2b2/stubs2015} & Clinical anonymization    & 47 & 45,443 / 5,439 / 32,587 \\
 & \corpusIdNerTwitter \cite{corpus/wnut/strauss-etal-2016-results} & Twitter posts    & 21 & 2,394 / 1,000 / 3,850 \\
 & \corpusIdNerSocial \cite{corpus/wnut/derczynski-etal-2017-results} & Social media   & 13 & 3.394 / 1,009 / 1.287 \\
 & \corpusIdNerWetlab \cite{corpus/wnut/tabassum2020} & Wetlab protocols               & 37 & 8.444 / 2,862 / 2,813 \\
 & \corpusIdNerLitbank \cite{corpus/litbank/bamman-etal-2019-annotated} & Literature   & 13 & 5.549 / 1.388 / 2.973 \\
 & \corpusIdNerSEC \cite{corpus/sec/salinas-alvarado-etal-2015-domain} & Financial     &  9 & 825 / 207 / 443 \\
 & \corpusIdNerSOFC \cite{friedrich-etal-2020-sofc} & Materials science                 &  9 & 490 / 123 / 263 \\ \midrule
\multirow{9}{*}{\rotatebox{90}{NER\& POS}}
 & \corpusIdGum \cite{corpus/zeldes-2017-gum} & All (GUM)      & 23/17 & 3,883 / 960 / 2,060 \\ 
 \cline{3-5}
 & \corpusIdGumAcademic  & Academic                            & 23/17 & 321 / 81 / 173 \\
 & \corpusIdGumBiography & Biography                           & 23/17 & 434 / 106 / 233 \\
 & \corpusIdGumFiction & Fiction                               & 23/17 & 576 / 144 / 309 \\
 & \corpusIdGumInterview & Interview                           & 23/17 & 599 / 150 / 321 \\
 & \corpusIdGumNews  & News                                    & 23/17 & 360 / 91 / 194 \\
 & \corpusIdGumReddit  & Reddit                                & 23/17 & 500 / 126 / 269 \\
 & \corpusIdGumTravel  & Travel                                & 23/17 & 431 / 108 / 232 \\
 & \corpusIdGumWikihow  & Wikihow                              & 23/17 & 612 / 154 / 329 \\ 
 \midrule
\multirow{3}{*}{\rotatebox{90}{POS}}
 & \corpusIdPosEWT \cite{corpus/ud-ewt/silveira14gold}& Blog. Email. Social                & 17 & 12,514 / 1,998 / 2.074 \\
 & \corpusIdPosLines \cite{corpus/ud-lines/ahrenberg-2015-converting} & (non-)Fiction. spoken          & 17 & 2,738 / 912 / 914 \\ 
 & \corpusIdPosPartut \cite{corpus/ud-partut/sanguinetti2014converting} & Legal. News. Wikipedia       & 17 & 1,781 / 156 / 153 \\ \midrule
\multirow{14}{*}{\rotatebox{90}{Temporal Expressions}}
 & \corpusIdTimeTimebank \cite{corpus/timebank/uzzaman-etal-2013-semeval} & News     & 9 & 2,557 / 640 / 303 \\
 & \corpusIdTimeAquaint \cite{corpus/timebank/uzzaman-etal-2013-semeval} & News      & 9 & 972 / 243 / 522 \\
 & \corpusIdTimeAncient \cite{corpus/strotgen-etal-2014-extending} & Historical Wikipedia    & 9 & 456 / 114 / 245 \\
 & \corpusIdTimeWikiwars \cite{corpus/mazur-dale-2010-wikiwars} & Wikipedia                 & 9 & 2,788 / 697 / 1,494 \\
 & \corpusIdTimeSMS \cite{corpus/strotgen-gert-2013-journal} & SMS                             & 9 & 1,674 / 419 / 898 \\
 & \corpusIdTimeSCI \cite{corpus/strotgen-gert-2013-journal} & Clinical                        & 9 & 461 / 116 / 248 \\
 & \corpusIdTimeClinical \cite{corpus/i2b2/sun2012} & Clinical                  & 9 & 5,943 / 1,486 / 5,665 \\  
 & \corpusIdAce \cite{corpus/walker-etal-2006-ace} & All (ACE-05)    & 9 & 8,958 / 2,241 / 4,802 \\
 \cline{3-5}
 & \corpusIdAceConversations  & Broadcast conversations       & 9 & 1,655 / 414 / 887 \\
 & \corpusIdAceNews  & Broadcast news                          & 9 & 2,087 / 522 / 1,119 \\
 & \corpusIdAceTelephony  & Conversational telephony           & 9 & 1,756 / 440 / 942 \\
 & \corpusIdAceNewswire  & Newswire                            & 9 & 1,172 / 293 / 628 \\
 & \corpusIdAceUsenet  & Usenet                                & 9 & 1,168 / 292 / 626 \\
 & \corpusIdAceWebblog  & Webblog                              & 9 & 1,120 / 280 / 600 \\
 \bottomrule
\end{tabular}
\caption{Overview of dataset domains and their sizes used in the transfer experiments.}
\label{tab:datasets}
\end{table*}


\begin{table*}[bhp!]
  \footnotesize
  \centering
  
  \begin{subtable}[h]{0.33\textwidth}
  \centering
  \begin{tabular}{l|llb} \toprule
  Corpus & \multicolumn{3}{c}{BERT} \\
  & Pre & Rec & \fscore \\ \midrule 
   \corpusIdNerCONLL            & 90.5 & 91.9 & 91.2 \\ 
   \corpusIdNerWetlab           & 78.2 & 81.0 & 79.6 \\
   \corpusIdNerSocial           & 60.1 & 35.9 & 44.9 \\
   \corpusIdNerTwitter          & 46.8 & 44.7 & 45.7 \\ 
   \corpusIdNerClinical         & 82.0 & 85.8 & 83.9 \\ 
   \corpusIdNerAnonymization    & 94.7 & 93.2 & 94.0 \\ 
   \corpusIdNerSEC              & 76.7 & 87.9 & 81.9 \\
   \corpusIdNerLitbank          & 66.1 & 74.5 & 70.0 \\
   \corpusIdNerSOFC             & 73.3 & 82.8 & 77.8 \\ 
   \corpusIdGumAcademic         & 46.3 & 58.8 & 51.8 \\
   \corpusIdGumBiography        & 61.0 & 72.1 & 66.1 \\
   \corpusIdGumFiction          & 62.8 & 72.0 & 67.1 \\ 
   \corpusIdGumInterview        & 48.9 & 58.7 & 53.4 \\
   \corpusIdGumNews             & 43.7 & 52.7 & 47.8 \\ 
   \corpusIdGumReddit           & 50.5 & 61.9 & 55.6 \\ 
   \corpusIdGumTravel           & 37.7 & 51.0 & 43.3 \\
   \corpusIdGumWikihow          & 40.0 & 49.0 & 44.0 \\
   \corpusIdGum                 & 55.1 & 64.3 & 59.4 \\ 
  \bottomrule
  \end{tabular}
  \caption{Named Entity Recognition}
  \end{subtable}
  \hfill
  \begin{subtable}[h]{0.33\textwidth}
  \centering
  \begin{tabular}{l|llb} \toprule
  Corpus & \multicolumn{3}{c}{BERT} \\
  & Pre & Rec & \fscore \\ \midrule
   \corpusIdTimeTimebank        & 75.2 & 76.3 & 75.7 \\
   \corpusIdTimeAquaint         & 77.6 & 77.6 & 77.6 \\
   \corpusIdTimeAncient         & 71.8 & 79.6 & 75.5 \\
   \corpusIdTimeWikiwars        & 87.1 & 90.7 & 88.9 \\ 
   \corpusIdTimeSMS             & 63.8 & 68.4 & 66.0 \\
   \corpusIdTimeSCI             & 55.9 & 51.6 & 53.7 \\ 
   \corpusIdTimeClinical        & 72.2 & 76.7 & 74.4 \\ 
   \corpusIdAceConversations    & 60.5 & 64.0 & 62.2 \\
   \corpusIdAceNews             & 60.3 & 71.5 & 65.4 \\ 
   \corpusIdAceTelephony        & 39.0 & 55.6 & 45.8 \\
   \corpusIdAceNewswire         & 76.8 & 81.9 & 79.2 \\ 
   \corpusIdAceUsenet           & 56.5 & 65.2 & 60.5 \\ 
   \corpusIdAceWebblog          & 65.6 & 69.4 & 67.5 \\ 
   \corpusIdAce                 & 66.9 & 77.6 & 71.8 \\ 
  \bottomrule
  \end{tabular}
  \caption{Temporal Expression Extraction}
  \end{subtable}
  \hfill
  \begin{subtable}[h]{0.27\textwidth}
  \centering
  \begin{tabular}{l|b} \toprule
  Corpus & \multicolumn{1}{c}{BERT} \\
  & \fscore=Acc. \\ \midrule 
   \corpusIdPosPartut           & 96.9 \\ 
   \corpusIdPosEWT              & 97.0 \\
   \corpusIdPosLines            & 97.5 \\
   \corpusIdGumAcademic         & 95.1 \\ 
   \corpusIdGumBiography        & 96.3 \\ 
   \corpusIdGumFiction          & 96.8 \\
   \corpusIdGumInterview        & 95.5 \\
   \corpusIdGumNews             & 95.9 \\ 
   \corpusIdGumReddit           & 94.6 \\ 
   \corpusIdGumTravel           & 94.5 \\
   \corpusIdGumWikihow          & 94.9 \\
   \corpusIdGum                 & 96.5 \\
  \bottomrule
  \end{tabular}
  \caption{POS Tagging}
  \end{subtable}
  
  \caption{Single task learning performance.}
  \label{tab:stl-performance}
\end{table*}

\setlength\tabcolsep{5pt}
\begin{table*}
\footnotesize
\centering
\begin{tabular}{l|lll|ll|ll} \toprule
Task & Min. & Avg. & Max. & \# Pos. & Pos. Avg. & \# Neg. & Neg. Avg. 
\\ \midrule
\multicolumn{4}{l}{\textit{Zero-Shot Model Transfer}} \\
\enspace NER  & -57.3  (-37.9) & -17.7 (-10.1) & 18.1 (8.0) 
& 7 /64 & 8.6 (3.8) & 56 /64 & -21.3 (-12.0) 
\\
\enspace POS  & -8.7   (-8.4)  & -2.8  (-2.7)  &  1.6 (1.5) 
& 13 /144 & 0.9 (0.8) & 127 /144 & -3.3 (-3.2) 
  \\
\enspace TIME & -100.0 (-83.2) & -42.7 (-29.6) & 38.6 (17.7) 
& 13 /196 & 10.3 (5.6) & 183 /196 & -46.5 (-31.7) 
\\
\multicolumn{8}{l}{\textit{Supervised Domain Adaptation}} \\
\enspace NER  & -5.2 (-2.7)   & 3.8 (1.9) & 14.5 (6.3) 
& 55 /64 & 4.7 (2.3) & 8 /64 & -1.7 (-0.9)
  \\
\enspace POS  & -0.3 (-0.3)   & 0.4 (0.4) & 1.8  (1.7)
& 116 /144 & 0.5 (0.5) & 9 /144 & -0.1 (0.1) 
  \\
\enspace TIME & -15.3 (-10.1) & 3.4 (2.0) & 32.7 (15.1)
& 133 /196 & 6.0 (3.7) & 62 /196 & -2.1 (-1.4) 
  \\
\multicolumn{8}{l}{\textit{Cross-Task Transfer}} \\
\enspace NER$\rightarrow$NER  & -9.1 (-4.1) & -0.2 (-0.2) & 6.8 (3.1)
& 39 /90 & 1.1 (0.6) & 46 /90 & -1.4 (-0.9) 
  \\
\enspace POS$\rightarrow$NER  & -5.9 (-4.8) & -0.5 (-0.3) & 2.6 (1.2)
& 42 /120 & 2.6 (1.2) & 65 /120 & -1.3 (-0.9) 
  \\
\enspace TIME$\rightarrow$NER & -7.2 (-3.3) & -0.9 (-0.5) & 0.9 (0.6) 
& 35 /150 & 0.4 (0.3) & 100 /150  & -1.5 (-0.9) 
  \\ \midrule
\end{tabular}
\caption{Transfer gains (\fscore increase) and the number of positive and negative transfer scenarios for each setting.
}
\label{tab:gains-single-task-long}
\end{table*}
\setlength\tabcolsep{6pt}

\setlength\tabcolsep{3.3pt}
\begin{table*}
\centering
\footnotesize
\begin{tabular}{l|cc|cc|cc|cc|cc|cc|cc|cc|cc|cc} \toprule
& \multicolumn{6}{c|}{\textit{Zero-Shot Model Transfer}}
& \multicolumn{6}{c|}{\textit{Supervised Domain Adaptation}}
& \multicolumn{6}{c|}{\textit{Cross-Task Transfer}} 
& \multicolumn{2}{c}{\textit{Average}} \\
& \multicolumn{2}{c|}{NER} & \multicolumn{2}{c|}{POS} & \multicolumn{2}{c|}{TIME} 
& \multicolumn{2}{c|}{NER} & \multicolumn{2}{c|}{POS} & \multicolumn{2}{c|}{TIME} 
& \multicolumn{2}{c|}{\begin{tabular}[c]{@{}l@{}}NER\\$\rightarrow$NER\end{tabular}} 
& \multicolumn{2}{c|}{\begin{tabular}[c]{@{}l@{}}POS\\$\rightarrow$NER\end{tabular}}
& \multicolumn{2}{c|}{\begin{tabular}[c]{@{}l@{}}TIME\\$\rightarrow$NER\end{tabular}}
& & \\
Distance & \avgRank & N
& \avgRank & N 
& \avgRank & N
& \avgRank & N 
& \avgRank & N 
& \avgRank & N
& \avgRank & N
& \avgRank & N 
& \avgRank & N 
& \avgRank & N 
\\ \midrule
Vocabulary 
    & 3.1 & 90.8 & 1.3 & 91.7 & 2.9 & \best{93.9} 
    & 3.6 & 83.7 & 2.3 & 92.4 & \best{2.6} & 90.7 
    & 5.3 & 82.8 & 5.4 & \best{87.7} & 8.4 & 84.3 
    & 3.9 & 88.7 \\
Annotation 
    & 3.5 & 89.7 & 1.3 & 91.7 & \best{2.4} & 93.8
    & 3.9 & 85.1 & 2.3 & 92.4 & 3.1 & 90.5 
    & 5.3 & 86.2 & \best{5.2} & \best{87.7} & 7.9 & 84.0
    & 3.9 & \best{89.1} \\
Datasize 
    & 5.0 & 81.0 & 2.1 & 88.3 & 3.6 & 90.0 
    & 4.6 & 81.2 & 2.8 & 89.9 & 4.1 & 86.7 
    & 6.9 & 75.0 & 5.5 & 87.3 & 9.1 & 84.6 
    & 4.9 & 84.9\\
Term Dist. 
    & 4.4 & 86.1 & \best{1.0} & \best{92.8} & 2.9 & 92.5 
    & 5.0 & 83.4 & 3.5 & 90.4 & 4.1 & 88.6 
    & 4.9 & 87.1 & 6.3 & 85.0 & 8.8 & 83.5 
    & 4.5 & 87.7 \\
LM Perp. 
    & 4.9 & 83.8 & 2.1 & 91.2 & 4.6 & 81.8
    & 3.9 & \best{86.8} & \best{1.8} & \best{93.5} & 4.6 & 84.4 
    & 4.8 & 84.8 & 6.8 & 86.5 & 6.1 & 81.9 
    & 4.4 & 86.1 \\ 
Text Emb. 
    & 3.5 & 88.8 & 4.0 & 86.1 & 4.4 & 89.4 
    & 4.0 & 84.3 & 5.0 & 84.1 & 4.9 & 86.5 
    & 4.7 & 84.3 & 9.6 & 84.7 & 7.0 & \best{84.8}  
    & 5.2 & 85.9\\
Task Emb. 
    & 3.4 & 89.4 & 4.0 & 86.3 & 5.0 & 89.8 
    & 3.8 & 83.7 & 5.0 & 84.3 & 5.2 & 86.3 
    & 4.9 & 84.5 & 7.2 & 82.8 & 7.8 & 83.3 
    & 5.1 & 85.6\\
Model Sim. 
    & \best{2.6} & \best{91.8} & 2.7 & 88.7 & 3.2 & 91.9 
    & \best{3.5} & 85.9 & 3.3 & 88.4 & 3.1 & \best{91.8}
    & \best{3.0} & \best{87.2} & 6.6 & 84.5 & \best{5.7} & 84.6
    & \best{3.7} & 88.3\\ \bottomrule

\end{tabular}
\caption{Ranking results for different similarity measures in the three transfer settings. The values displayed are the average rank of the best model (\avgRank) and the NDCG-score (N) compared to the observed performance.}
\label{tab:predictions}
\end{table*}
\setlength\tabcolsep{6pt}

\section{Model Performance}
\label{sec:appendix:stl}
We list the performance for all single-task models in Table~\ref{tab:stl-performance} with precision, recall and micro-\fscore for NER and TIME corpora and accuracy for POS.

\section{Fine-grained Results}
In addition to Table \ref{tab:gains-single-task} and Table \ref{tab:predictions2} that display task-wise averages, we report more fine-grained results in Table \ref{tab:gains-single-task-long} and Table \ref{tab:predictions}.

\section{Similarity Measures}
This section provides a more detailed overview of the similarity measures introduced in Section~\ref{sec:similarity}. 

\textit{Target vocabulary overlap} is the percentage of unique words from the target corpus covered in the source corpus.
    In contrast to vocabulary overlap, this is an asymmetric measure.
\textit{Annotation overlap} is a special case considering only annotated words. 
    
We also experiment with the \textit{Language model perplexity} \cite{baldwin-etal-2013-noisy} between two datasets. 
For this, a language model, in our case a 5-gram LM with Kneser–Ney smoothing \cite{heafield-2011-kenlm} as used by \citet{dai-etal-2019-using}, is trained for each source domain and tested against the target domain. The resulting perplexity gives hints how similar these domains are, i.e., a lower perplexity indicates similarity between domains.
    
\textit{Jensen-Shannon divergence}  \cite{ruder-plank-2017-learning} compares the term distributions between two texts, which are probability distributions that capture the frequency of words.
It is similar to vocabulary overlap, as it describes the textual overlap, but based on distributions instead of sets of terms.
    
A \textit{Text embedding}  \cite{vu-etal-2020-exploring} can be computed by extracting the feature vectors of a neural model. For this, the output of the last layer is averaged over all words in the dataset. This vector then represents the text domain. The distance between two vectors is computed by using cosine similarity.
    
The \textit{Task embedding} \cite{vu-etal-2020-exploring} takes a labeled source dataset and computes a representation based on the Fisher Information Matrix, which captures the change of model parameters w.r.t.\ the computed loss. 
This method assumes that similar tasks require similar parameters changes.
We use the code released by \citet{vu-etal-2020-exploring} to compute task embeddings from the different components of our BERT models and similarly use reciprocal rank fusion \cite{cormack2009reciprocal} to combine these.

\end{document}